\pdfoutput=1

\documentclass[11pt]{article}
\usepackage{hyperref}
\usepackage[preprint]{acl}

\usepackage{times}
\usepackage{latexsym}

\usepackage[T1]{fontenc}

\usepackage[utf8]{inputenc}

\usepackage{microtype}

\usepackage{inconsolata}

\usepackage{graphicx}
\usepackage{subcaption}

\usepackage{float}
\title{HistoLens: An LLM-Powered Framework for Multi-Layered Analysis of Historical Texts - A Case Application of Yantie Lun}

\author{Yifan Zeng \\
  Sun Yat-sen University, Guangzhou, China \\
  \texttt{zengyf53@mail2.sysu.edu.cn} 
  }

\begin{document}
\maketitle
\begin{abstract}
This paper proposes HistoLens, a multi-layered analysis framework for historical texts based on Large Language Models (LLMs). Using the important Western Han dynasty text "Yantie Lun" as a case study, we demonstrate the framework's potential applications in historical research and education. HistoLens integrates NLP technology (especially LLMs), including named entity recognition, knowledge graph construction, and geographic information visualization. The paper showcases how HistoLens explores Western Han culture in "Yantie Lun" through multi-dimensional, visual, and quantitative methods, focusing particularly on the influence of Confucian and Legalist thoughts on political, economic, military, and ethnic. We also demonstrate how HistoLens constructs a machine teaching scenario using LLMs for explainable analysis, based on a dataset of Confucian and Legalist ideas extracted with LLM assistance. This approach offers novel and diverse perspectives for studying historical texts like "Yantie Lun" and provides new auxiliary tools for history education. The framework aims to equip historians and learners with LLM-assisted tools to facilitate in-depth, multi-layered analysis of historical texts and foster innovation in historical education.
\end{abstract}

\section{Introduction}
Traditional historical research often relies on qualitative methods that depend on the subjective judgments and experiences of scholars \cite{Qua} and requires substantial time for reading historical texts. To address these challenges, this study proposes an interdisciplinary analysis framework, HistoLens, which combines natural language processing technologies, including Large Language Models (LLMs), to integrate qualitative and quantitative methods in historical research \cite{Qua}. This approach not only allows for drawing objective and verifiable conclusions but also enables rapid processing of large volumes of historical text data, presenting knowledge through diverse charts and graphs to offer new insights and perspectives \cite{DuanNature}.

LLMs, which are trained on vast amounts of textual data, demonstrate powerful capabilities in understanding and generating text across various languages \cite{LLMsTrain}, including classical Chinese \cite{Classical1,Class2,Classical3}. The use of LLMs to assist in the analysis of historical texts has become a novel and effective approach for historians. With well-crafted prompts \cite{Prom1,Prom2}, historians can adeptly utilize digital tools for rapid and multifaceted textual analysis. The LLM systems utilized in this study include Claude 3.5 Sonnet \cite{anthropic_claude_3_5}, GPT-4o mini \cite{openai_gpt_4o_mini}, as well as moonshot-v1\footnote{\url{https://kimi.moonshot.cn/}} and ERNIE 3.5\footnote{\url{https://yiyan.baidu.com/}}.

The LLM-powered historical text analysis framework includes the following main steps:

\noindent\textbf{Step 1. thematic word frequency analysis}: Tokenization, removal of stop words, and statistical analysis of key thematic word frequencies and their distribution in the original text.


\noindent\textbf{Step 2. LLM-based named entity and relationship recognition}: Identifying named entities such as individuals and places and extracting relationships between them by using LLM.

\noindent\textbf{Step 3. knowledge graph construction}: Building knowledge graphs based on entity and relationship information extracted by LLM.

\noindent \textbf{Step 4. spatiotemporal analysis}: Combining Geographic Information System (GIS) technology to visualize geographical distributions and time series in historical texts.

\noindent \textbf{Step 5. ideological school analysis}: Constructing text datasets containing historical thoughts, annotated with LLM assistance.

\noindent \textbf{Step 6. machine teaching scenario}: Based on the above ideological text datasets, using LLMs' text classification explanation capabilities to teach learners key features of thoughts in historical texts.

In this paper, we use "Yantie Lun" \cite{Translate} as an example to demonstrate the application of HistoLens in analyzing Confucian and Legalist ideological cultures, which to a certain extent determined the politics, economy, and military affairs of the Western Han period. "Yantie Lun" which records the Salt and Iron Conference held in 81 BCE \cite{ConLegAgainst}, is a crucial historical source for understanding the politics, economy, and ideologies of the mid-Western Han period. Through our proposed HistoLens, we explore and verify a range of cultural and historical viewpoints and phenomena in a multi-layered, visual, and quantitative manner, and offer new perspectives and vitality to the study of Western Han history, and construct a novel Confucianism and Legalism dataset while making a novel machine teaching scenario via explainable analysis. Details of the Salt and Iron Conference, Legalism and Confucianism are shown in appendix. The code and data are available.

\begin{figure}[h]
  \centering
  \begin{subfigure}[b]{0.47\columnwidth}
    \includegraphics[width=0.95\columnwidth]{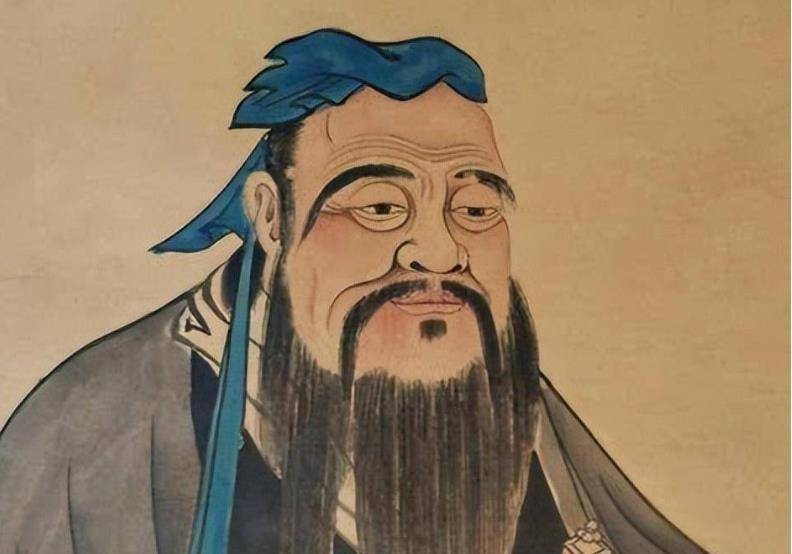}
    \caption{}
    \label{fig:sub1}
  \end{subfigure}
  \begin{subfigure}[b]{0.45\columnwidth}
    \includegraphics[width=0.95\columnwidth]{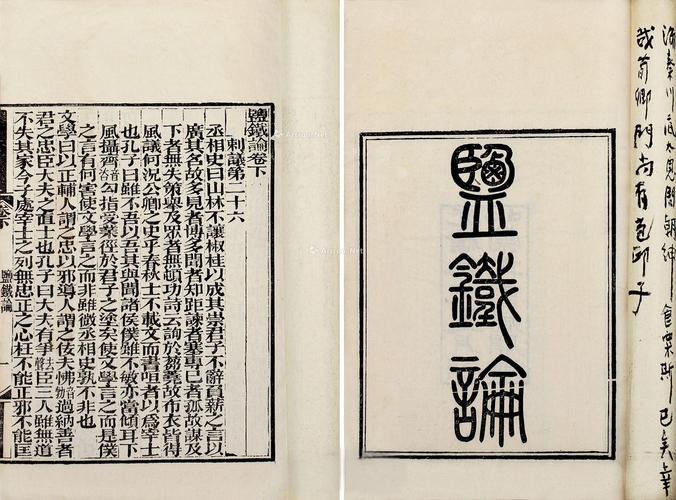}
    \caption{}
    \label{fig:sub2}
  \end{subfigure}
  \vspace{-10pt}
  \caption{(a) 
Confucius. (b) A photo of the Yantie Lun.}
  \label{fig:test}
\end{figure}

HistoLens is not limited to analyzing "Yantie Lun" but provides a generalized, universal LLM-assisted analysis framework for the study and learning of other historical texts. HistoLens offers new tools and methods for historical research and learning, enabling scholars to process large volumes of historical documents more rapidly and comprehensively, discovering new historical patterns and insights through multi-layered analysis.

\section{Proposed Framework}

\subsection{Thematic word frequency analysis}
Word frequency analysis is a commonly used paradigm in cliometrics \cite{WordFre}. Through word frequency analysis, we identified key terms of six themes: Confucianism, Legalism, agriculture, economy, Huaxia region, and ethnic minorities. Confucian terms outweigh Legalist ones, as Sang also employs Confucian concepts in debate. This could indicate a synthesis of both two schools, reflecting "outer Confucianism, inner Legalism" \cite{RUBiaofali}.

\begin{figure}[H]
  \centering
  \begin{subfigure}[b]{0.49\columnwidth}
    \includegraphics[width=\columnwidth]{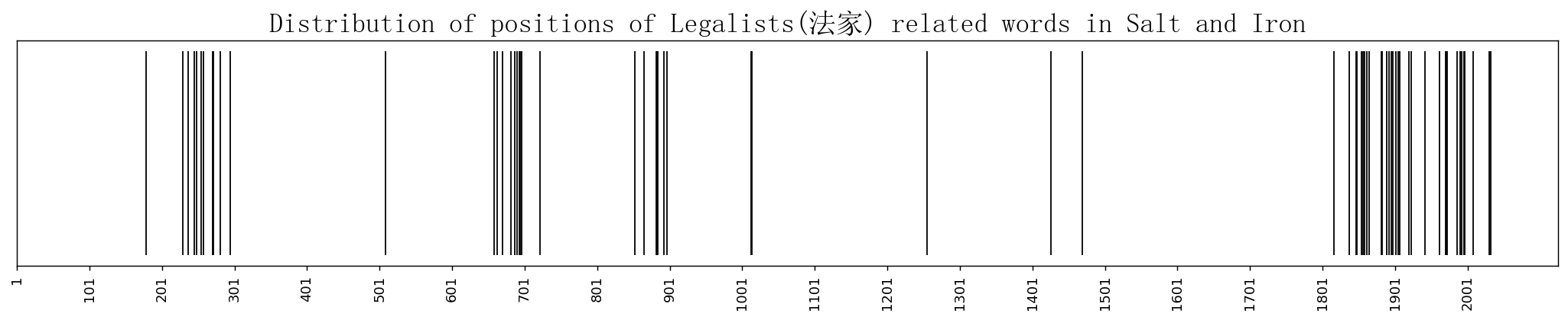}
    \caption{Legalism}
    \label{fig:sub1}
  \end{subfigure}
  \hfill 
  \begin{subfigure}[b]{0.49\columnwidth}
    \includegraphics[width=\columnwidth]{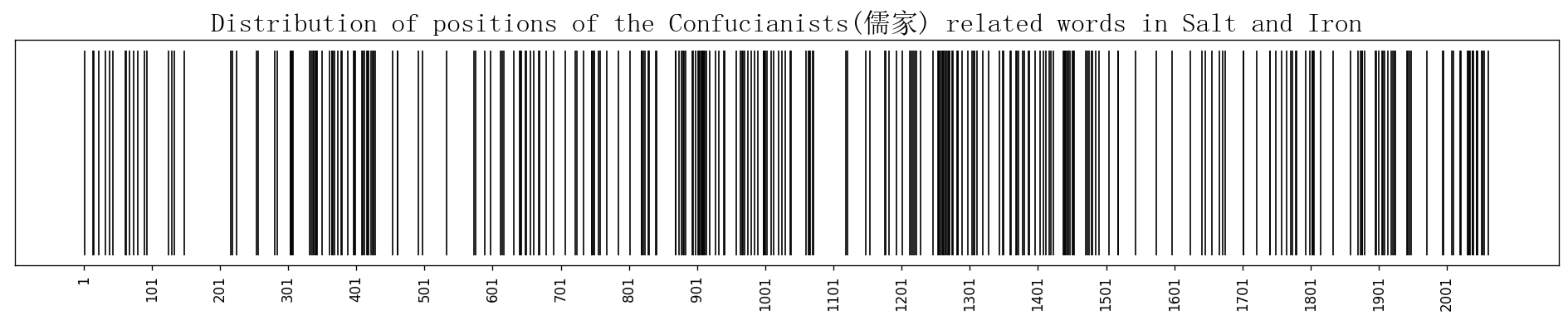}
    \caption{Confucianism}
    \label{fig:sub2}
  \end{subfigure}
    \begin{subfigure}[b]{0.49\columnwidth}
    \includegraphics[width=\columnwidth]{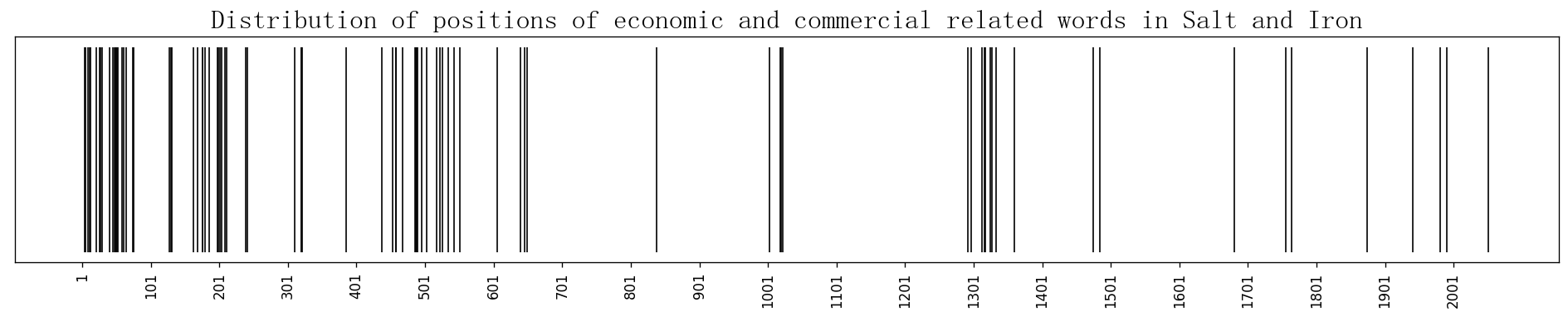}
    \caption{Economy}
    \label{fig:sub2}
  \end{subfigure}
  \begin{subfigure}[b]{0.49\columnwidth}
    \includegraphics[width=\columnwidth]{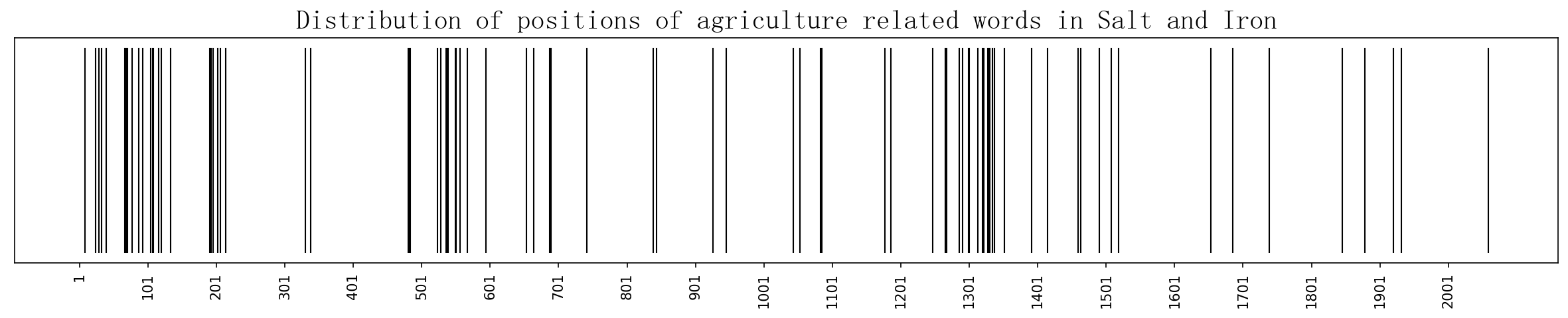}
    \caption{Agriculture}
    \label{fig:sub2}
  \end{subfigure}
  \begin{subfigure}[b]{0.49\columnwidth}
    \includegraphics[width=\columnwidth]{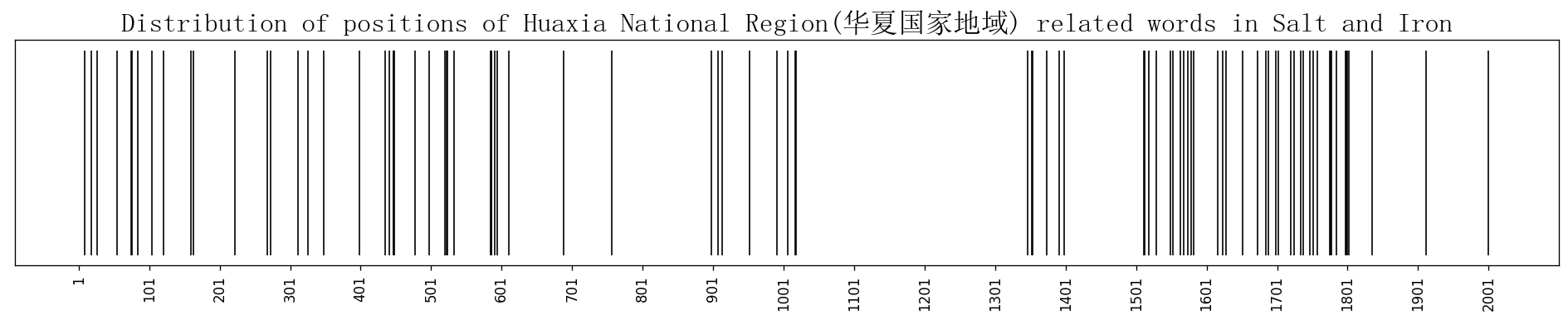}
    \caption{Huaxia region}
    \label{fig:sub2}
  \end{subfigure}
      \begin{subfigure}[b]{0.49\columnwidth}
    \includegraphics[width=\columnwidth]{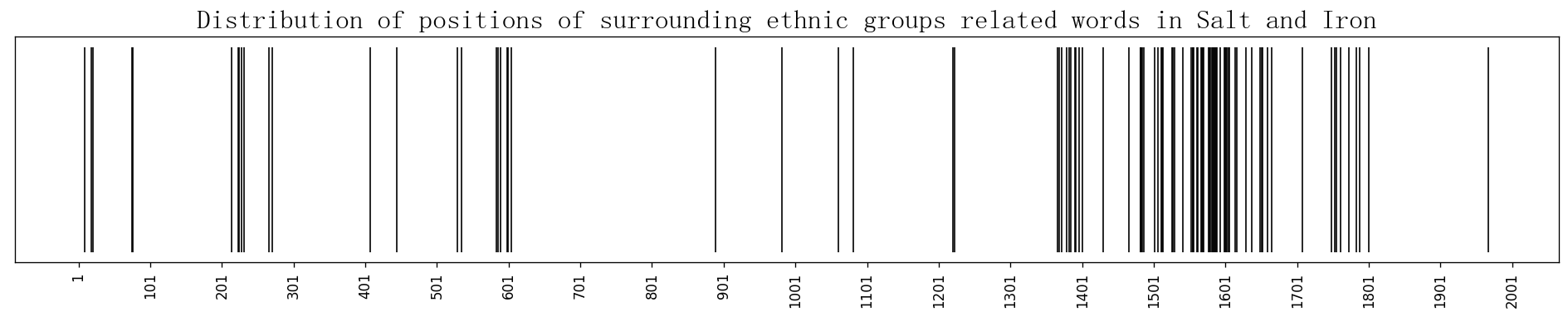}
    \caption{Ethnic minorities}
    \label{fig:sub2}
  \end{subfigure}
  \caption{The distributions of topic keywords in text.
}
  \label{fig:test}
\end{figure}
\vspace{-15pt}

\begin{figure}[h]
  \centering
  \begin{subfigure}[b]{1\columnwidth}
    \includegraphics[width=\columnwidth]{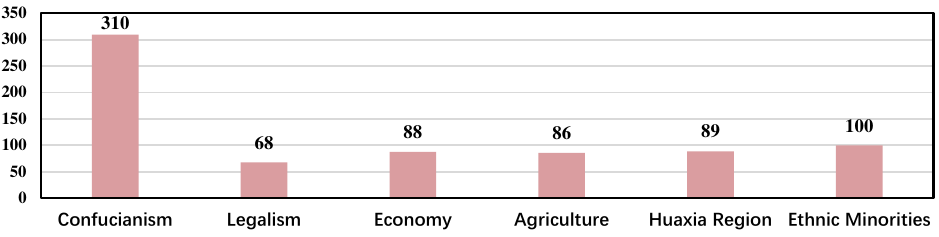}
  \end{subfigure}
  \vspace{-20pt}
  \caption{Bar chart of the number of lines.}
\end{figure}
\vspace{-10pt}

They emphasized Han Dynasty's agriculture, state-controlled trade and commerce. It also discussed foreign relations, wars with the Xiongnu, and early Chinese nationalism. The language used showed a cultural identity linked to the Han people and their perception of other ethnic groups.


\subsection{LLM named entity recognition}
LLMs can effectively extract named entities such as individuals and places \cite{NER}. This allows knowledge in historical text to be represented through knowledge graphs \cite{KnowGraph} and GIS \cite{GIS}.
\vspace{-12pt}
\begin{figure}[H]
  \centering
  \begin{subfigure}[b]{1\columnwidth}
    \includegraphics[width=\columnwidth]{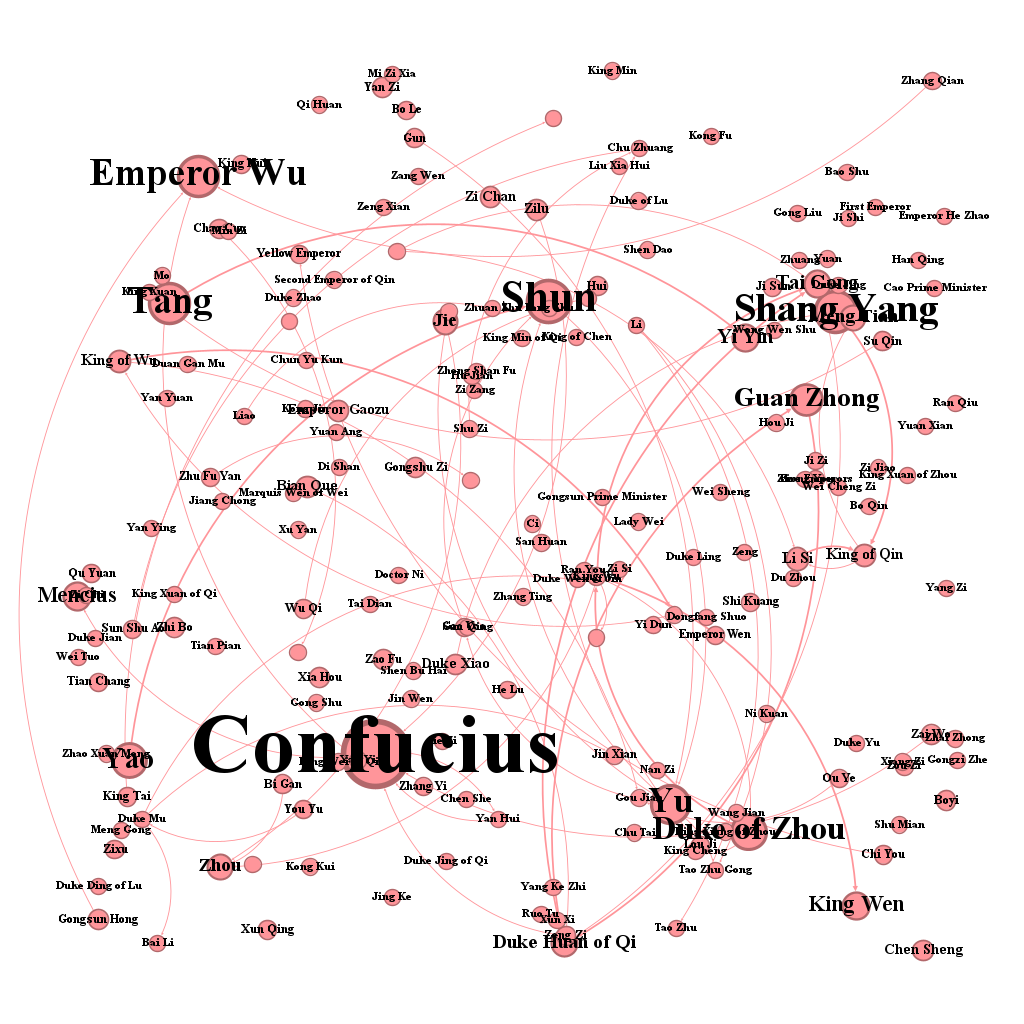}
    \label{fig:sub2}
  \end{subfigure}
  \vspace{-40pt}
  \caption{The knowledge graph of the figures mentioned in Yantie Lun.}
  \label{fig:test}
\end{figure}

\vspace{-20pt}

\begin{figure}[H]
  \centering
  \begin{subfigure}[b]{1\columnwidth}
    \includegraphics[width=\columnwidth]{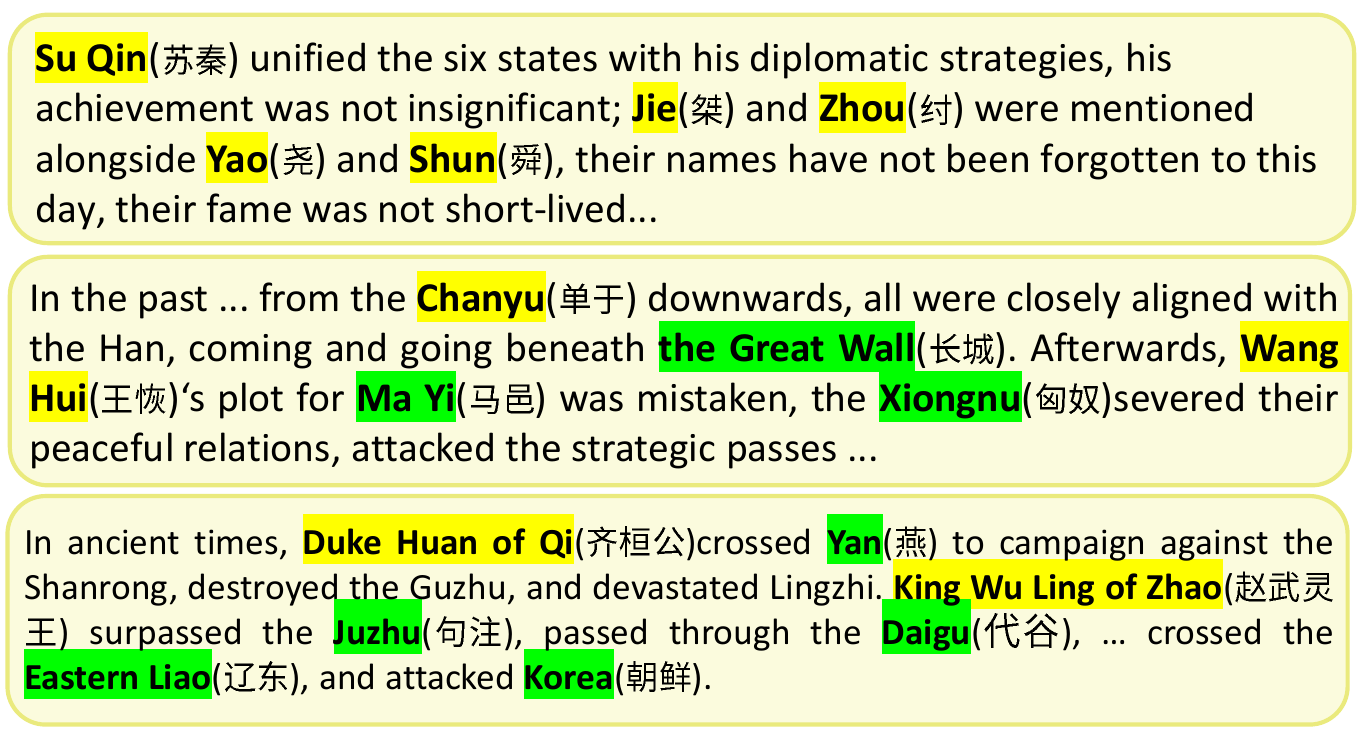}
    \label{fig:sub2}
  \end{subfigure}
  \vspace{-40pt}
  \caption{LLMs extract entities of figures and places (entities of people and places are represented in yellow and green, respectively).}
  \label{fig:test}\end{figure}

  \vspace{-10pt}
We used LLMs to extract historical figures from the Yantie Lun, noting their frequency and relationships to created a graph using Gephi. Confucius, as the Confucian figurehead, was a central point of xianliang wenxue's discussion. Shang Yang, representing Legalism, was both praised by Sang and criticized by Confucians, becoming a focal point of disagreement. After Dong Zhongshu's promotion of Confucianism, figures such as Confucius, Mencius, King Wen, Yao, Shun, and Yu, who serve as symbols of Confucianism, received positive mentions from both sides. Even Sang cited Confucian exemplars, such as Shun, Yu and Tang. This indicates some convergence of Legalism and Confucian thought \cite{Convergen}. Confucian ideas like the golden age of the Three Dynasties, became mainstream or political correctness in mid-Western Han. Moreover, Qin Shihuang and Shang Yang were frequently mentioned, highlighting the lessons drawn from the Qin Empire, which firstly achieved unification but quickly collapsed \cite{Qin}.

\begin{figure}[h]
  \centering
  \begin{subfigure}[b]{1\columnwidth}
    \includegraphics[width=\columnwidth]{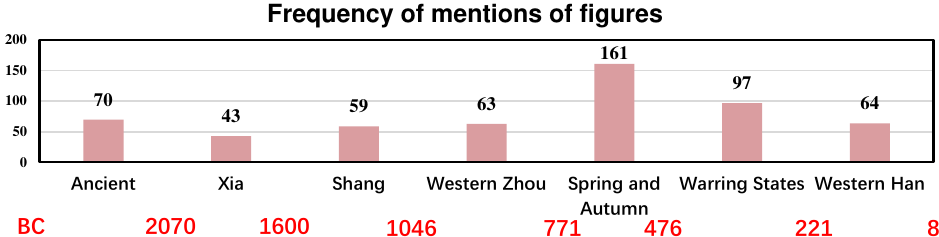}
  \end{subfigure}
  \vspace{-20pt}
  \caption{Historical period figure mention frequency. }
  \label{fig:FigFreqbar}
\end{figure}

\vspace{-5pt}
Further analysis of figures mentioned more than twice by era (shown in Figure~\ref{fig:FigFreqbar}) reveals a tendency to cite ancient examples. This reflects a widespread tendency among Western Han intellectuals to look back to the past, venerate antiquity, and hold a historical view that favors the past over the present ("ancient" appears 121 times in the full text), which later influenced Wang Mang's extensive restoration of ancient systems \cite{WangMang}. This can also verify the Axial Age theory proposed by the Jaspers \cite{Axis} and show that the Spring and Autumn philosophical diversity marked a creative peak in Chinese civilization.

We used LLMs to extract place name entities from the Yantie Lun and plotted they on a map using a geographic information system (GIS). This allows us to analyze the geographical and spatial understandings of the Western Han intellectual elite during that period, including their perception of the world layout and the relationship between the Huaxia region and surrounding ethnic minorities.

\begin{figure}[h]
  \centering
  \begin{subfigure}[b]{0.9\columnwidth}
    \includegraphics[width=\columnwidth]{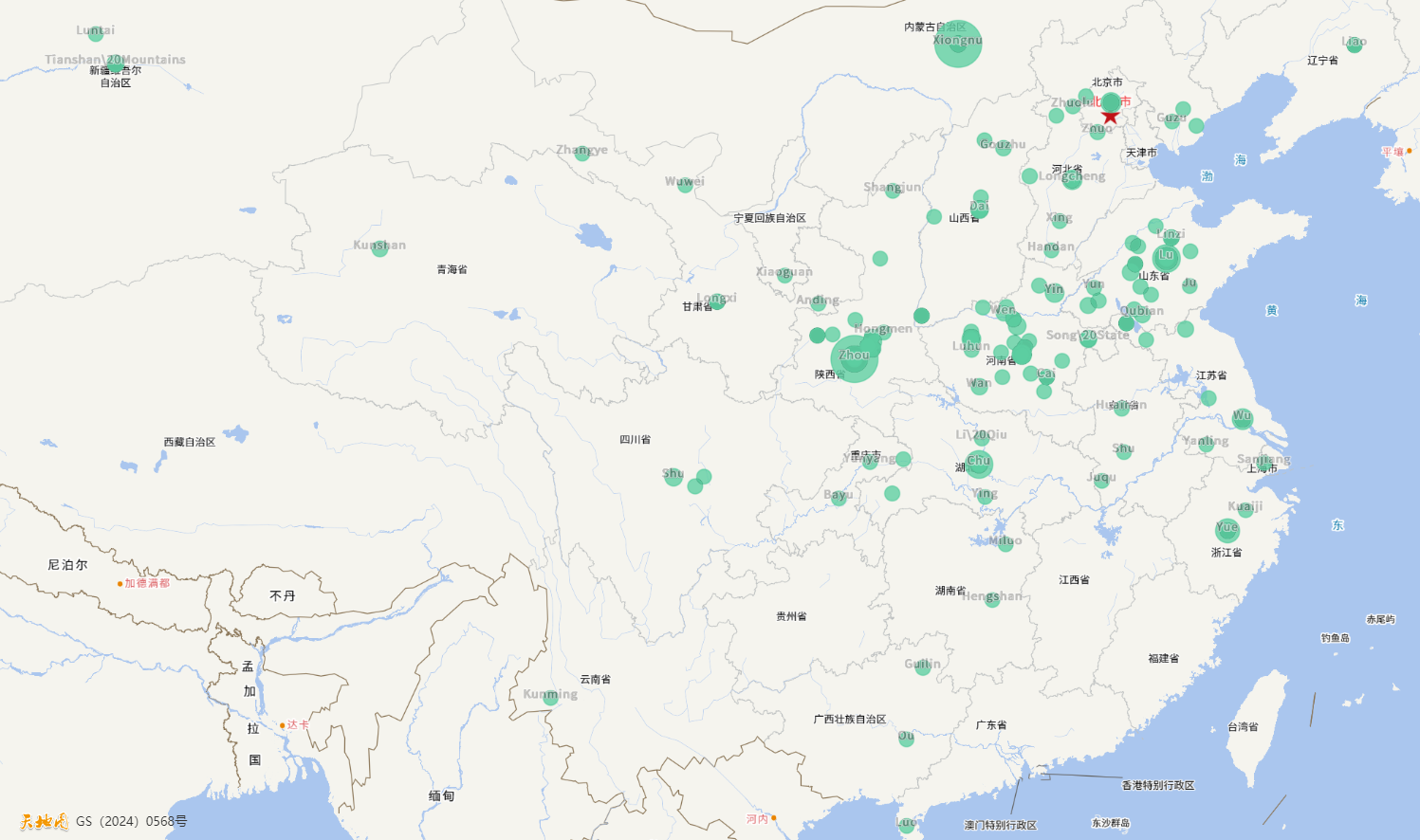}
    \label{fig:sub2}
  \end{subfigure}
  \vspace{-20pt}
  \caption{The map of the places mentioned in text.}
  \label{fig:test}
\end{figure}

The text discusses the historical concentration of place names along the Yellow River, emphasizing the region's cultural and economic importance during China's early dynasties, such as Xia, Shang, and Han. It contrasts this with the later southward shift of China's centers during the Jin, Tang, and Song dynasties \cite{South}. Notable areas include ancient states like Jin, Zheng, and Yan, as well as the Qi and Lu regions where Confucianism began. The Xiongnu's Great Desert region and the Hexi Corridor, which led to the Western Regions and was pivotal for the Silk Road under Emperor Wu, are also highlighted \cite{Hexi}. Additionally, the text mentions the Yue people in the southeast, indicating their significance in Qin and Han times \cite{Baiyue}.

\subsection{Confucianism and Legalism dataset}
Utilizing deep learning and natural language processing (NLP) algorithms, we aim to gain a deeper understanding of the Confucianism and Legalism ideologies in the Yantie Lun. By processing the text, extracting debates, and labeling passages that represent Confucian or Legalism views, we construct a dataset of these philosophical perspectives. The dataset can be utilized for various tasks such as text classification, sentiment analysis, creating educational tools, and text generation using NLP algorithms. Such an approach offers new perspectives and tools for studying and researching ancient Chinese philosophy and history of Western Han.
\begin{figure}[h]
  \centering
  \begin{subfigure}[b]{0.9\columnwidth}
    \includegraphics[width=\columnwidth]{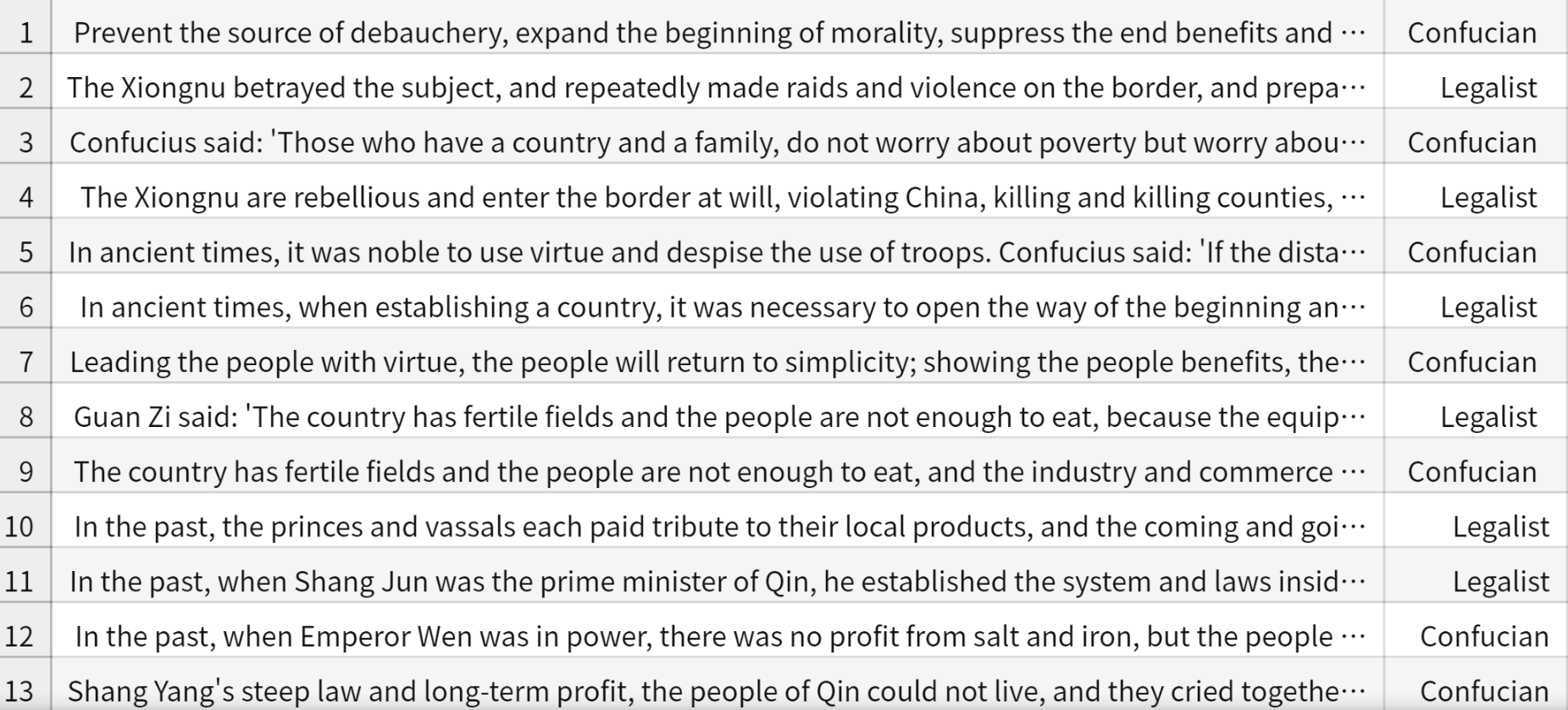}
    \label{fig:sub2}
  \end{subfigure}
  \vspace{-20pt}
  \caption{Part of Confucianism and Legalism dataset.}
  \label{fig:test}
  \vspace{-14pt}
\end{figure}
The dataset contains 116 statements representing Legalism views (from Sang Hongyang) and 123 representing Confucianism (from xianliang wenxue), totaling 239. Each statement contains an average of 188 Chinese characters, amounting to approximately 45,000 characters in total. The number of statements from both sides is roughly equal, showing balanced arguments. It can be seen that the author Huan Kuan, a Confucian scholar, did not show particular bias towards the Confucianism side due to his educational background. And this can avoid algorithmic bias caused by imbalance. The dataset contains complete sentences to avoid taking things out of context. The dataset has removed the identities of speakers to prevent the algorithm from guessing based on the speaker's identity. Data is in classical Chinese, with some rare characters used in ancient times, which poses difficulties for humans who have not received professional training, and also presents a challenge for NLP algorithms.

\subsection{Interpretability analysis and machine teaching}

Humanities scholars without a background in computer science can achieve surprisingly good results by using simple LLM prompting techniques to process historical texts. Numerous studies and practices have shown that some prompting tricks can make LLMs perform even beyond expectations. In a novel approach, we have employed prompting engineering to classify and analyze the interpretability of the dataset proposed in the previous section. The prompting engineering techniques we used include role-playing (having the model act as an expert), chain-of-thought (CoT) \cite{CoT}, and few-shot learning \cite{fewshotGPT3}.

Interpretability analysis \cite{Explain} enables models to provide the basis for text classification, allowing humans to understand the internal mechanisms behind the model's decision-making. This not only fosters trust in the decisions made by the model (or enables the rapid identification of errors in the model's decision basis), but also facilitates AI models in teaching humans how to capture the decisive and key characteristics and differences between Confucian and Legalist thought. It achieves machine teaching while performing machine learning. Analysis can teach beginners that a text belongs to Confucianism as it contains terms like "benevolence", "morality" and "education", which are typical characteristics of Confucian thought. If machine learning involves humans giving vast amounts of data for models to learn from, machine teaching imparts knowledge acquired by models to humans (for example, helping beginners easily grasp the distinctions between Confucianism and Legalism). It can also offer humanities scholars novel perspectives after learning a large corpus of texts, presenting an innovative approach to learning and research.
\begin{figure}[h]
  \centering
  \begin{subfigure}[b]{1\columnwidth}
    \includegraphics[width=\columnwidth]{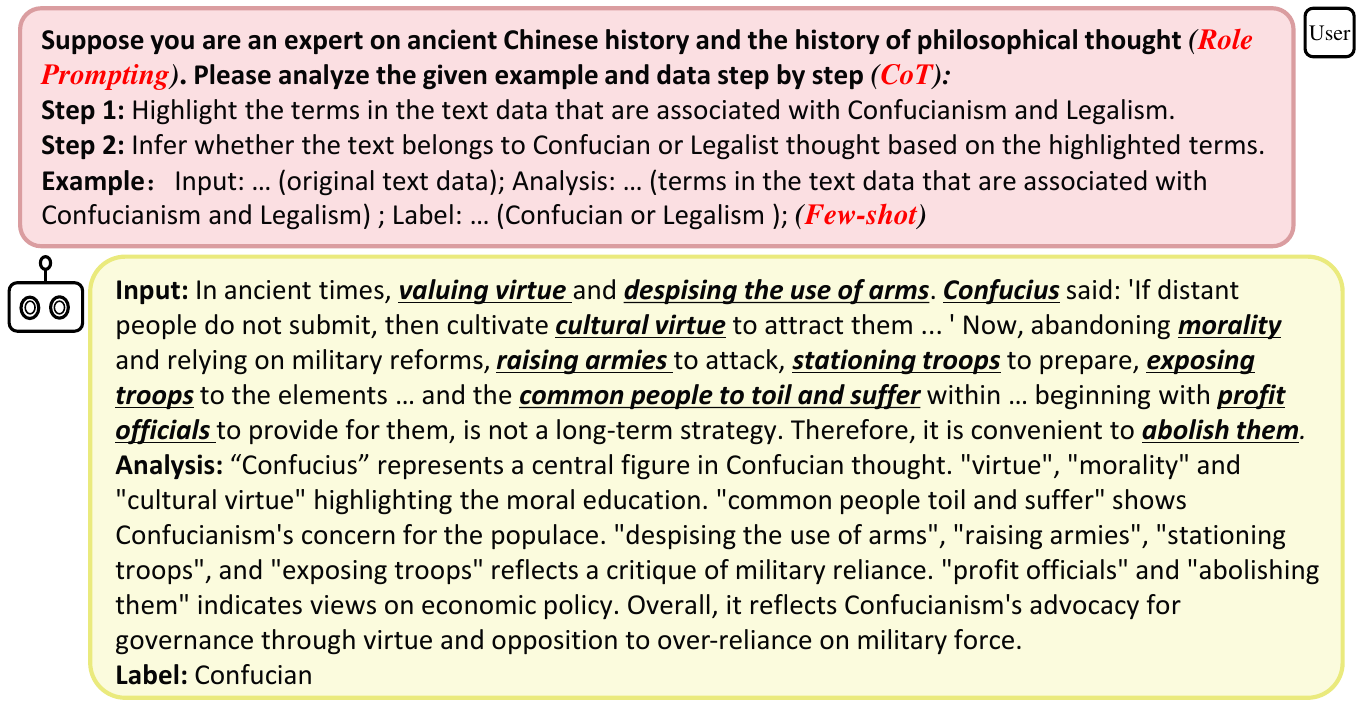}
    \label{fig:sub2}
  \end{subfigure}
  \vspace{-25pt}
  \caption{Applying LLM prompt engineering for text classification and interpretability analysis on the dataset.}
  \label{fig:test}
\end{figure}
\vspace{-15pt}

\section{Conclusion}
This paper proposes HistoLens, an LLM-powered framework for multi-layered analysis of historical texts, demonstrated through a case of the Western Han text "Yantie Lun".  HistoLens offers a novel approach to historical research and education and goes beyond traditional qualitative analysis, providing a comprehensive exploration of historical time, space, figures, and ideologies. Through HistoLens, we've investigated key ideological features of the Western Han, such as the debate and integration of Confucianism and Legalism (as "outer Confucianism, inner Legalism"), historical decline mindset, the impact of the Axial Age, and the Huaxia national consciousness. HistoLens offers new tools for historians and learners to efficiently analyze historical documents. It not only offers fresh insights into "Yantie Lun", but also presents a generalizable approach for analyzing other historical texts across various periods and cultures, and uncover insights through multi-layered analysis. In conclusion, HistoLens represents a step forward in the application of AI to historical research and education.

\section*{Limitations}

While HistoLens offers numerous advantages, it's important to acknowledge its limitations and areas for future improvement:

\noindent \textbf{LLM Limitations:} The effectiveness and accuracy of LLM on classical historical texts might be limited due to lack of specialized fine-tuning. Future work could involve developing or fine-tuning LLM specifically for historical texts' ancient language processing, attempting to further mitigate LLM's hallucinations regarding historical facts.

\noindent \textbf{Scope of Application:} Due to resource constraints, this study primarily applied HistoLens on "Yantie Lun," which represents a specific historical period (Chinese Western Han Dynasty). Expanding the application of HistoLens to a broader range of historical texts from different periods and cultures, such as Plutarch's "Parallel Lives" from the Roman period, would enhance its generalizability.

\noindent \textbf{Comprehensiveness:} While our proposed HistoLens's analysis covers multiple dimensions, such as historical figures and their relationships, locations and ideologies, it may not encompass all relevant historical phenomena, such as genre analysis of historical works from different periods. Further interdisciplinary collaboration could help improve HistoLens to identify and incorporate additional relevant factors.

\noindent \textbf{Validation:} The accuracy of LLM-assisted analysis needs rigorous validation against traditional historical research methods. Developing standardized evaluation metrics for computational historical analysis would be beneficial.

\noindent \textbf{Ethical Considerations:} The use of AI in historical interpretation raises questions about potential biases, such as implicit value judgments in historical events, which may be Eurocentric or reflect great-power chauvinism, potentially offending certain national sentiments. Future work should address these ethical implications and develop guidelines for responsible use of AI in historical research and learning.

\noindent \textbf{User Interface and Accessibility:} While not addressed in this paper, developing a user-friendly interface for HistoLens would be crucial for its widespread adoption among historians and educators who may not have extensive computer technical expertise.

\noindent \textbf{Integration with Multimodal Historical Data:} Despite language being the primary form of historical narration, future iterations of HistoLens could leveraging Multimodal Large Language Model (MLLM) to enhance its analytical capabilities by integrating multimodal historical data. This includes the vast amount of historical photographs and even videos that have been preserved since the invention of photography.

\section{Statements}
We used LLM systems for academic research purposes and provided appropriate citations. We cited the sources appropriately when using "Yantie Lun" for academic research.

\bibliography{custom}

\begin{thebibliography}{30}
\providecommand{\natexlab}[1]{#1}

\bibitem[{{Anthropic}(2024)}]{anthropic_claude_3_5}
{Anthropic}. 2024.
\newblock {Claude 3.5 Sonnet}.
\newblock {\url{https://www.anthropic.com/news/claude-3-5-sonnet}}.
\newblock [Online; accessed 2024-10-03].

\bibitem[{Belinkov et~al.(2020)Belinkov, Gehrmann, and Pavlick}]{Explain}
Yonatan Belinkov, Sebastian Gehrmann, and Ellie Pavlick. 2020.
\newblock Interpretability and analysis in neural nlp.
\newblock In \emph{Proceedings of the 58th annual meeting of the association for computational linguistics: tutorial abstracts}, pages 1--5.

\bibitem[{Bogdanov et~al.(2024)Bogdanov, Constantin, Bernard, Crabb{\'e}, and Bernard}]{NER}
Sergei Bogdanov, Alexandre Constantin, Timoth{\'e}e Bernard, Benoit Crabb{\'e}, and Etienne Bernard. 2024.
\newblock Nuner: Entity recognition encoder pre-training via llm-annotated data.
\newblock \emph{arXiv preprint arXiv:2402.15343}.

\bibitem[{Brown(2020)}]{fewshotGPT3}
Tom~B Brown. 2020.
\newblock Language models are few-shot learners.
\newblock \emph{arXiv preprint arXiv:2005.14165}.

\bibitem[{Chen et~al.(2024)Chen, Lou, Chen, Bai, Xiang, Yang, Zhao, and Zhang}]{Class2}
Andong Chen, Lianzhang Lou, Kehai Chen, Xuefeng Bai, Yang Xiang, Muyun Yang, Tiejun Zhao, and Min Zhang. 2024.
\newblock Benchmarking llms for translating classical chinese poetry: Evaluating adequacy, fluency, and elegance.
\newblock \emph{arXiv preprint arXiv:2408.09945}.

\bibitem[{D{\'e}murger et~al.(2002)D{\'e}murger, Sachs, Woo, Bao, Chang, and Mellinger}]{South}
Sylvie D{\'e}murger, Jeffrey~D Sachs, Wing~Thye Woo, Shuming Bao, Gene Chang, and Andrew Mellinger. 2002.
\newblock Geography, economic policy, and regional development in china.
\newblock \emph{Asian Economic Papers}, 1(1):146--197.

\bibitem[{Duan et~al.(2023)Duan, Wang, Yang, and Su}]{DuanNature}
Siyu Duan, Jun Wang, Hao Yang, and Qi~Su. 2023.
\newblock Disentangling the cultural evolution of ancient china: a digital humanities perspective.
\newblock \emph{Humanities and Social Sciences Communications}, 10(1):1--15.

\bibitem[{Feng et~al.(2024)Feng, Zhang, Gu, Ye, He, and Wang}]{CoT}
Guhao Feng, Bohang Zhang, Yuntian Gu, Haotian Ye, Di~He, and Liwei Wang. 2024.
\newblock Towards revealing the mystery behind chain of thought: a theoretical perspective.
\newblock \emph{Advances in Neural Information Processing Systems}, 36.

\bibitem[{Feng(2023)}]{Qua}
Zhiwei Feng. 2023.
\newblock The four levels of digital humanities research.
\newblock \emph{Journal of Nanjing Normal University (Social Sciences Edition)}, (3):1--9.
\newblock In Chinese.

\bibitem[{Gale(2022)}]{Translate}
Esson~McDowell Gale. 2022.
\newblock \emph{Discourses on Salt and Iron: A Debate on State Control of Commerce and Industry in Ancient China: Chapters I-XIX translated from the chinese of Huan K'uan with introduction and notes}, volume~2.
\newblock Brill.

\bibitem[{Gieseking(2018)}]{GIS}
Jen~Jack Gieseking. 2018.
\newblock Where are we? the method of mapping with gis in digital humanities.
\newblock \emph{American Quarterly}, 70(3):641--648.

\bibitem[{Grajzl and Murrell(2024)}]{WordFre}
Peter Grajzl and Peter Murrell. 2024.
\newblock How machine learning will change cliometrics.
\newblock In \emph{Handbook of Cliometrics}, pages 2721--2750. Springer.

\bibitem[{Lin(2018)}]{HuoGuang}
Tsung-shun Lin. 2018.
\newblock New interpretation on the discussion on salt and iron--the political intention of huo guang and the unperceived situation of sang hong-yang.
\newblock \emph{Humanitas Taiwanica}, (89).

\bibitem[{Litaina et~al.(2024)Litaina, Soularidis, Bouchouras, Kotis, and Kavakli}]{LLMsTrain}
Tania Litaina, Andreas Soularidis, Georgios Bouchouras, Konstantinos Kotis, and Evangelia Kavakli. 2024.
\newblock Towards llm-based semantic analysis of historical legal documents.

\bibitem[{Liu(2024)}]{Convergen}
Dongwang Liu. 2024.
\newblock Differences and integration of political thought between ancient chinese confucianism and legalism.
\newblock \emph{Trans/Form/A{\c{c}}{\~a}o}, 47(4):e0240042.

\bibitem[{Liu(2021)}]{ConLegAgainst}
Qunyi Liu. 2021.
\newblock Yantie lun in the pro-legalist and anti-confucian campaign.
\newblock In \emph{European and Chinese Histories of Economic Thought}, pages 202--214. Routledge.

\bibitem[{Meacham(1996)}]{Baiyue}
William Meacham. 1996.
\newblock Defining the hundred yue.
\newblock \emph{Bulletin of the Indo-Pacific Prehistory Association}, 15:93--100.

\bibitem[{{OpenAI}(2024)}]{openai_gpt_4o_mini}
{OpenAI}. 2024.
\newblock {GPT-4O Mini: Advancing Cost-Efficient Intelligence}.
\newblock {\hyperlink{https://openai.com/index/gpt-4o-mini-advancing-cost-efficient-intelligence/}{https://openai.com/index/gpt-4o-mini-advancing-cost-efficient-intelligence/}}.
\newblock Accessed: 2024-10-4.

\bibitem[{Ouellette(2010)}]{Qin}
Patrick Ouellette. 2010.
\newblock \emph{Power in the Qin dynasty: Legalism and external influence over the decisions and legacy of the first emperor of China}.
\newblock Ph.D. thesis.

\bibitem[{Ozdemir(2023)}]{Prom1}
Sinan Ozdemir. 2023.
\newblock \emph{Quick start guide to large language models: strategies and best practices for using ChatGPT and other LLMs}.
\newblock Addison-Wesley Professional.

\bibitem[{Qin and Eisner(2021)}]{Prom2}
Guanghui Qin and Jason Eisner. 2021.
\newblock Learning how to ask: Querying lms with mixtures of soft prompts.
\newblock \emph{arXiv preprint arXiv:2104.06599}.

\bibitem[{Schefold(2019)}]{Economy}
Bertram Schefold. 2019.
\newblock A western perspective on the yantie lun.
\newblock In \emph{The political economy of the Han dynasty and its legacy}, pages 153--174. Routledge.

\bibitem[{Shi(2007)}]{WangMang}
Feng Shi. 2007.
\newblock A study on wang mang’s jade tablet for the feng-shan ceremony of the xin dynasty.
\newblock \emph{Chinese Archaeology}, 7(1):163--169.

\bibitem[{Smith(2015)}]{Axis}
Andrew Smith. 2015.
\newblock Between facts and myth: Karl jaspers and the actuality of the axial age.
\newblock \emph{International Journal of Philosophy and Theology}, 76(4):315--334.

\bibitem[{Wan(2012)}]{ExpandWar}
Ming Wan. 2012.
\newblock Discourses on salt and iron: A first century bc chinese debate over the political economy of empire.
\newblock \emph{Journal of Chinese Political Science}, 17:143--163.

\bibitem[{Windhager et~al.(2024)Windhager, Salisu, Liem, and Mayr}]{KnowGraph}
Florian Windhager, Saminu Salisu, Johannes Liem, and Eva Mayr. 2024.
\newblock The knowledge graph as a data sculpture: Visualising arts and humanities data with maps, graphs, and sets over time.
\newblock \emph{Geographical Research in the Digital Humanities; Bielefeld University Press: Bielefeld, Germany}, pages 113--134.

\bibitem[{Yu et~al.(2024)Yu, Zang, Wang, Zhuang, and Gu}]{Classical1}
Chengyue Yu, Lei Zang, Jiaotuan Wang, Chenyi Zhuang, and Jinjie Gu. 2024.
\newblock Charpoet: A chinese classical poetry generation system based on token-free llm.
\newblock In \emph{Proceedings of the 62nd Annual Meeting of the Association for Computational Linguistics (Volume 3: System Demonstrations)}, pages 315--325.

\bibitem[{Zhang(2021)}]{Hexi}
Defang Zhang. 2021.
\newblock Using excavated slips to look at effective governance of the northern frontier during the han dynasty—the lelang commandery in han slips.
\newblock \emph{Bamboo and Silk}, 4(2):336--364.

\bibitem[{Zhao et~al.(2024)Zhao, Wang, and Wang}]{Classical3}
Cheng Zhao, Bin Wang, and Zhen Wang. 2024.
\newblock Understanding literary texts by llms: A case study of ancient chinese poetry.
\newblock \emph{arXiv preprint arXiv:2409.00060}.

\bibitem[{Zhou(2011)}]{RUBiaofali}
Haiwen Zhou. 2011.
\newblock Confucianism and the legalism: A model of the national strategy of governance in ancient china.
\newblock \emph{Frontiers of Economics in China}, 6(4):616--637.

\end{thebibliography}

\appendix
\section{Data Availability}
The open-sourced code and data can be found here: \href{https://anonymous.4open.science/r/YantieLunData-8E57/}{https://anonymous.4open.science/r/YantieLunData-8E57/} (anonymous git repository for double-blind review). The full text and book images of Yantie Lun can be found here: \href{http://www.xueheng.net/}{http://www.xueheng.net/}.

\section{Detail background of the Salt and Iron Conference}
\label{sec:appendix1}

"Yantie Lun" (Discourses on Salt and Iron) is an important political treatise and historical document from ancient China. It serves as a crucial historical source for understanding the politics, economy, and ideologies of the mid-Western Han period. The Yantie Lun compiled by Huan Kuan records the proceedings of a debate held in 81 BCE (the 6th year of Emperor Zhao's Shiyuan era), which is known as the Salt and Iron Conference\cite{ConLegAgainst}. The conference was organized under the influence of Huo Guang, and was presided over by Imperial Censor (yushi dafu) Sang Hongyang \cite{HuoGuang}. More than 60 scholars (xianliang wenxue) from various regions were invited to participate in the discussions\cite{ConLegAgainst}. The conference was a measure to broaden channels for public opinion.

The conference (debate) and Yantie Lun appears to discuss economic policies \cite{Economy} on salt and iron, as well as warfare against the Xiongnu\cite{ExpandWar}, but it is fundamentally a confrontation and clash between Legalist and Confucian ideologies and political philosophies represented by Sang Hongyang and the scholars (xianliang wenxue), respectively \cite{ConLegAgainst}. This is the aspect that this paper focuses on exploring and illustrating. 

At that time, Imperial Censor Sang Hongyang held significant financial and economic power. During Emperor Wu of Han's aggressive military campaigns against neighboring ethnic minorities, years of foreign wars led to a massive fiscal deficit\cite{ExpandWar}. Emperor Wu ordered Sang Hongyang to generate more revenue for the imperial treasury.

In response, Sang Hongyang implemented a series of political and economic measures. These included state monopolies on essential goods like salt and iron production, alcohol taxation (jiuque), equitable transportation tax (junshu), price stabilization (pingzhun), and wealth control and taxation of merchants (suanmingaomin). These policies amassed vast wealth to support the empire's enormous military expenditures.
However, these measures also exhausted the people's resources, making life unbearable for many. In his later years, Emperor Wu issued the "Luntai Edict of Self-Criticism", reflecting on his excessive use of the people's strength, a mistake reminiscent of the Qin dynasty's downfall \cite{HuoGuang}. 
Consequently, Huo Guang, a powerful minister, sought to restore the Han Empire's strength, which emphasized allowing the people and the country to recuperate \cite{HuoGuang}. Huo had risen to political prominence through his family connection to Emperor Wu's favored Consort Wei.
It was under Huo Guang's initiative that the Salt and Iron Conference was convened. His aim may have been to use the criticisms of the Confucian scholars (xianliang wenxue) gathered from among the people to publicly reflect the popular opposition to the salt and iron policies. This would undermine Sang Hongyang's political capital and allow Huo to wrest control of financial and economic power from Sang Hongyang \cite{HuoGuang}.

\section{Brief background of Confucianism and Legalism}
\label{sec:app2}

Confucianism, represented by Confucius, emphasizes moral cultivation, advocating for moral education to guide people towards goodness and social harmony.
Legalism, as represented by Han Fei and Shang Yang, believes in the inherent selfishness of human nature and advocates for strict laws and punishments to maintain order. It bolsters monarchical authority and uses legal systems to centralize power and improve state efficiency.

\section{About extracting place name entities}
We used LLMs to extract place name entities from the Yantie Lun, and mapped these ancient Chinese locations to their modern counterparts. Some required manual correction, such as "Kunshan" referring to the Kunlun Mountains, not the modern city in Jiangsu province today, and "Zhongshan" denoting the Zhao kingdom's city, not the Guangdong province city, and "Penglai" referring to a mythical offshore mountain, not Penglai in Shandong Province today. 

\section{Supplement figures}

\begin{figure}[h]
  \centering
  \begin{subfigure}[b]{1\columnwidth}
    \includegraphics[width=\columnwidth]{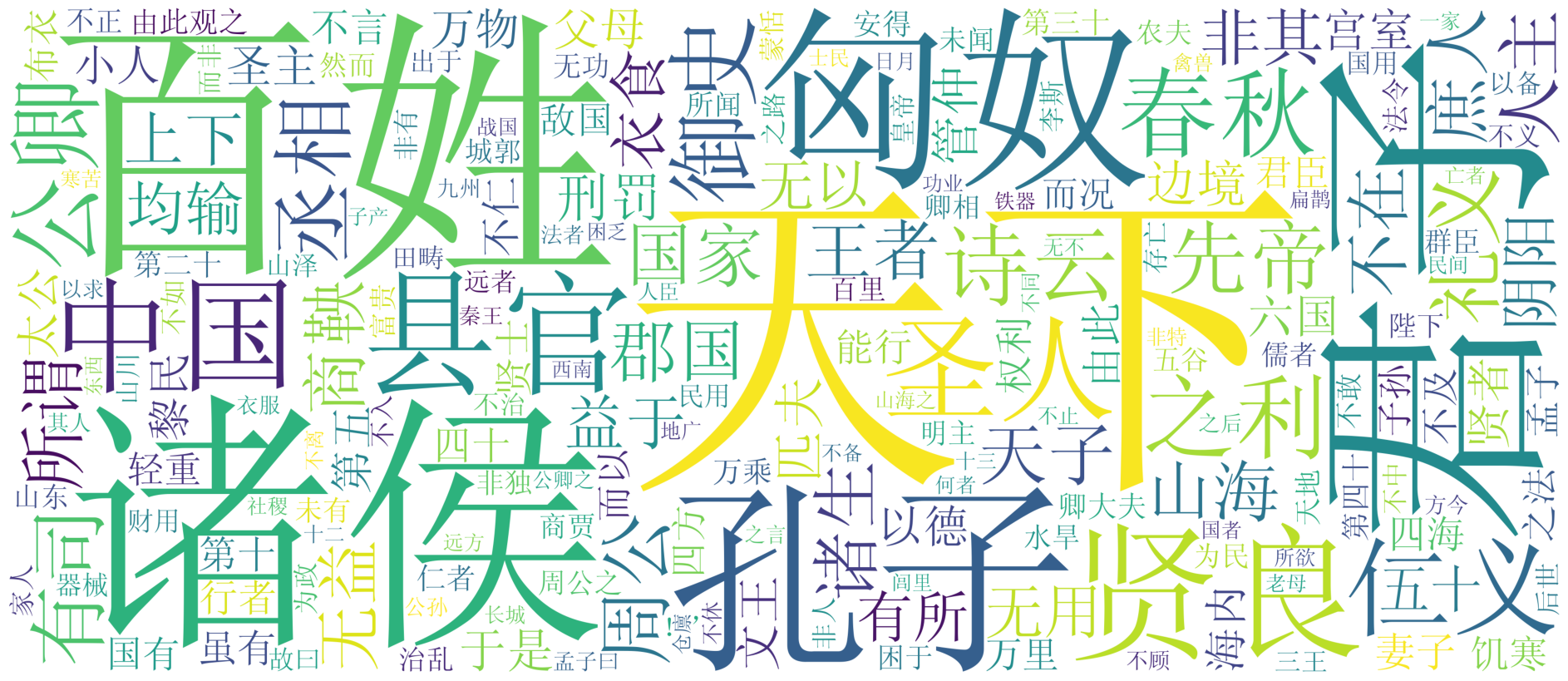}
  \end{subfigure}
  \vspace{-15pt}
  \caption{The word cloud of the Yantie Lun (Chinese version). Frequent terms like "gentleman (junzi)" "benevolence and righteousness (renyi)" "Confucius (kongzi)" and "common people (baixing)" are all closely related to Confucian doctrine. Yantie Lun contains a wealth of Confucian thought, demonstrating that Confucianism was already very popular in the mid-Western Han Dynasty, and the author Huan Kuan himself had a strong Confucian background.}
\end{figure}

\begin{figure}[h]
  \centering
  \begin{subfigure}[b]{0.9\columnwidth}
    \includegraphics[width=\columnwidth]{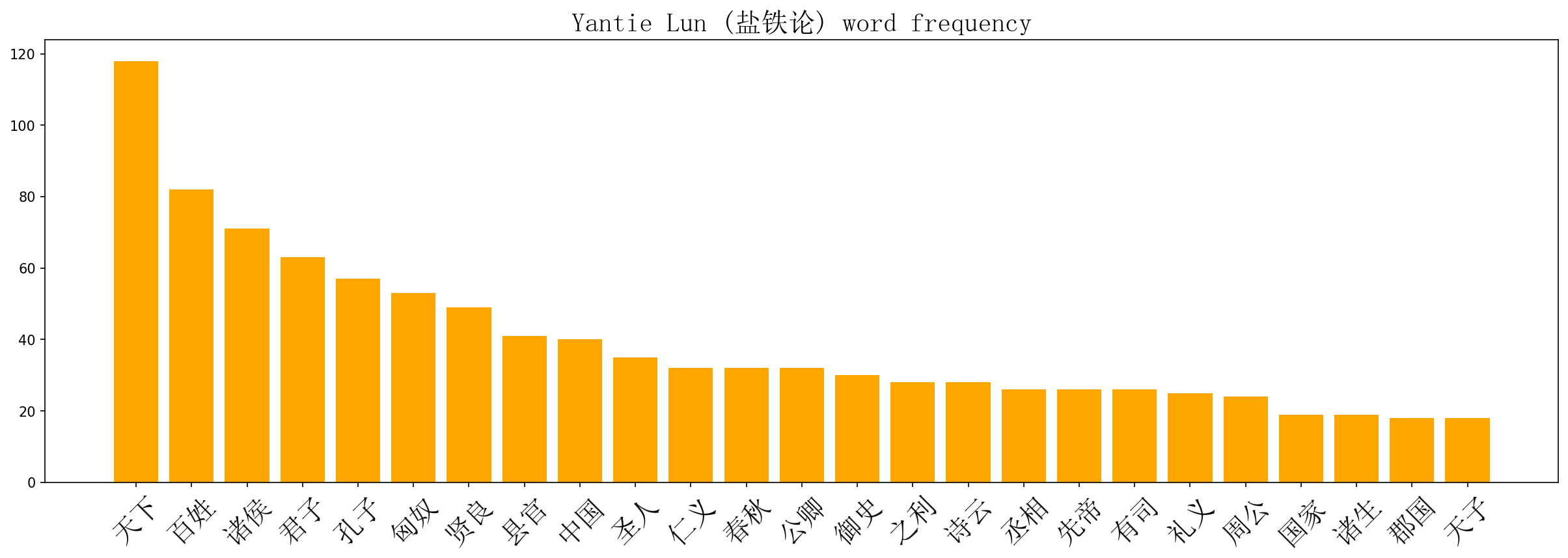}
  \end{subfigure}
  \vspace{-10pt}
  \caption{The word frequency of the Yantie Lun (Chinese version). 
}
\end{figure}

\begin{figure}[h]
  \centering
  \begin{subfigure}[b]{1\columnwidth}
    \includegraphics[width=\columnwidth]{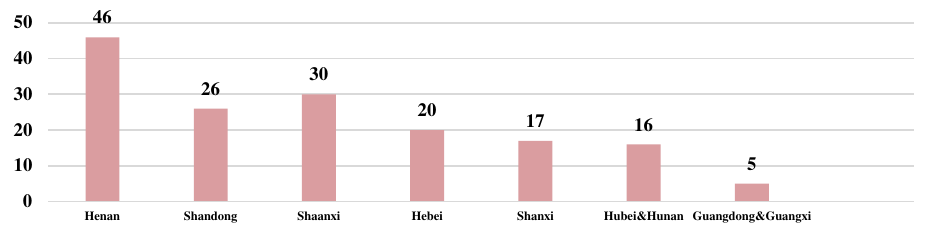}
  \end{subfigure}
  \vspace{-20pt}
  \caption{The geographical distribution of place names mentioned in the text.}
\end{figure}

\begin{figure}[h]
  \centering
  \begin{subfigure}[b]{1\columnwidth}
    \includegraphics[width=\columnwidth]{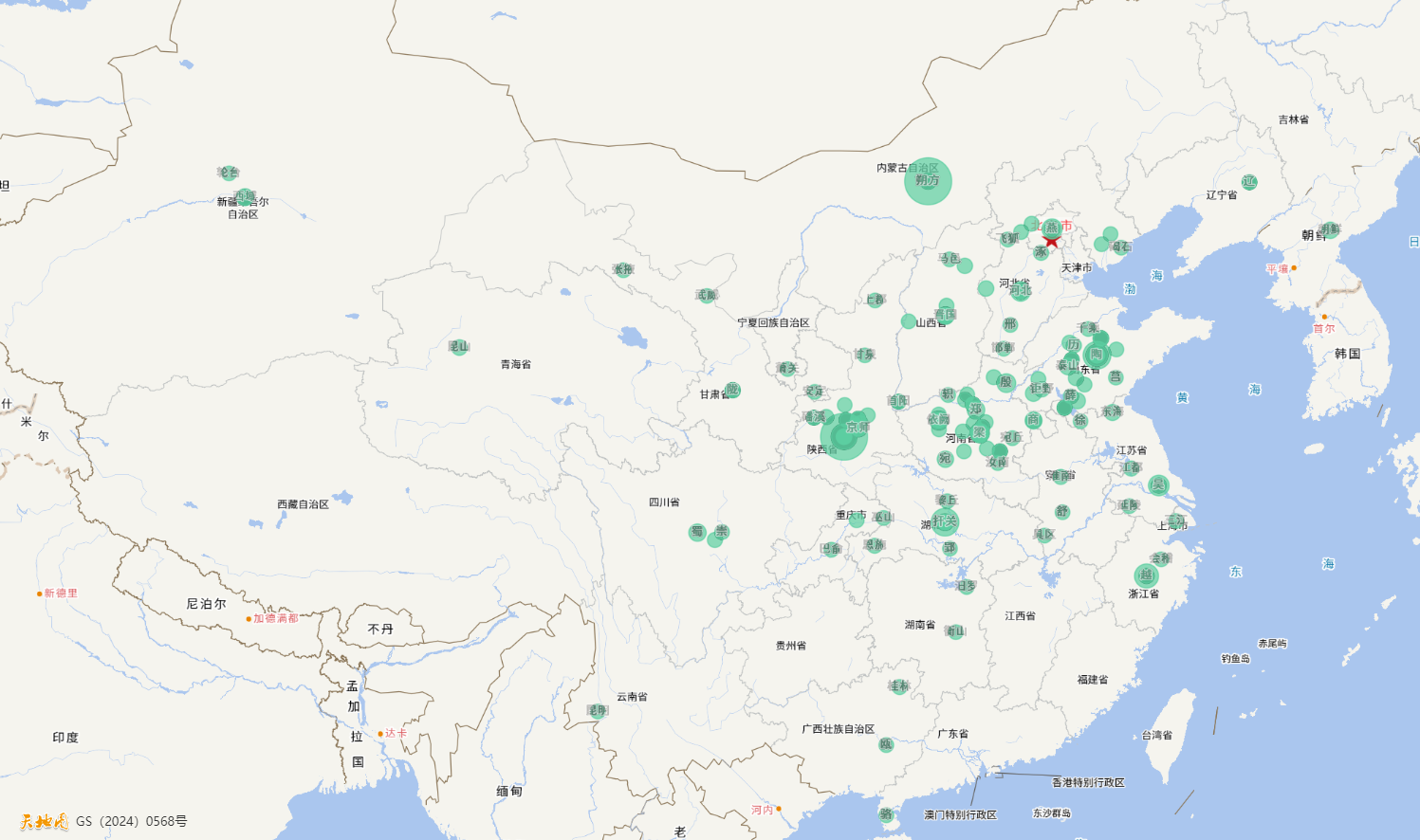}
  \end{subfigure}
  \vspace{-20pt}
  \caption{The map of places mentioned in the Yantie Lun (Chinese version).
}
\end{figure}

\begin{figure}[h]
  \centering
  \begin{subfigure}[b]{1\columnwidth}
    \includegraphics[width=\columnwidth]{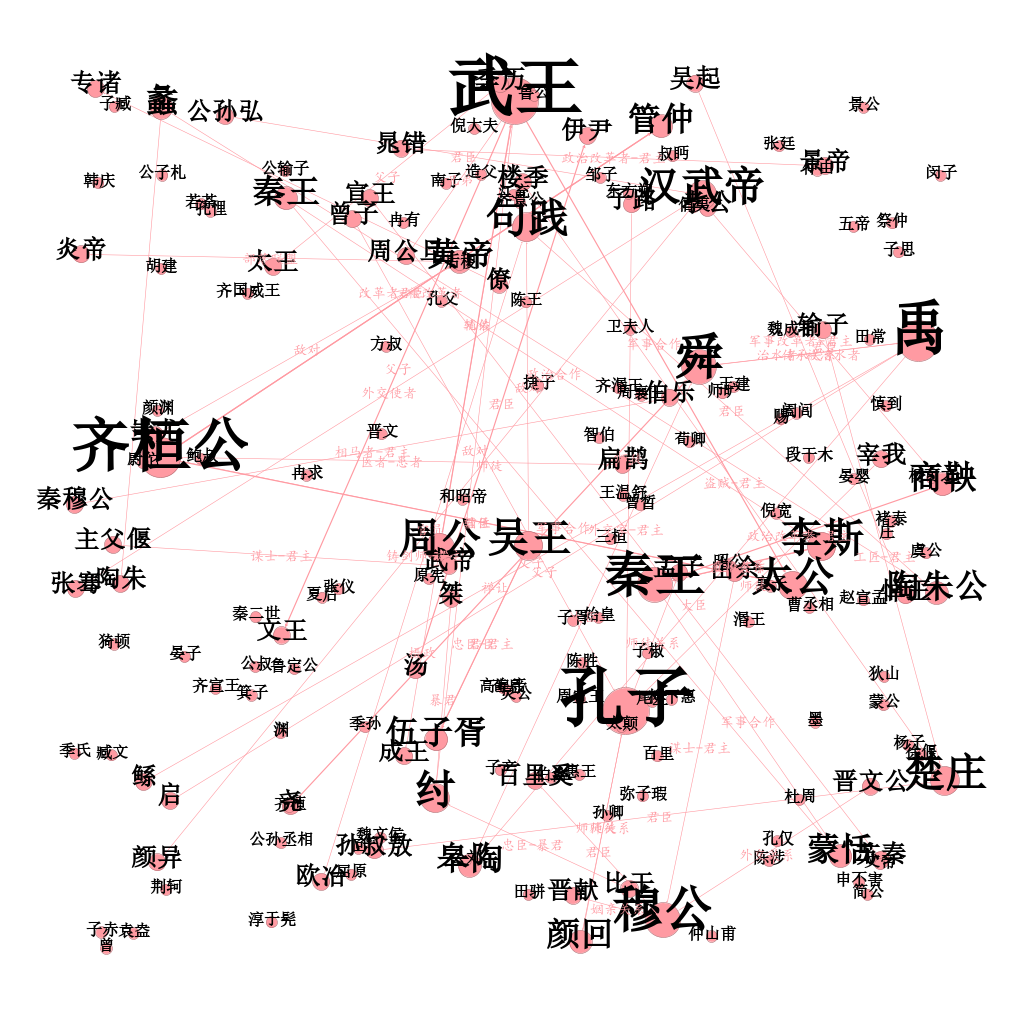}
    \label{fig:sub1}
  \end{subfigure}
  \vspace{-40pt}
  \caption{The knowledge graph of the figures mentioned in the Chinese version of Yantie Lun.
}
  \label{fig:test}
\end{figure}

\begin{figure}[h]
  \centering
  \begin{subfigure}[b]{1\columnwidth}
    \includegraphics[width=\columnwidth]{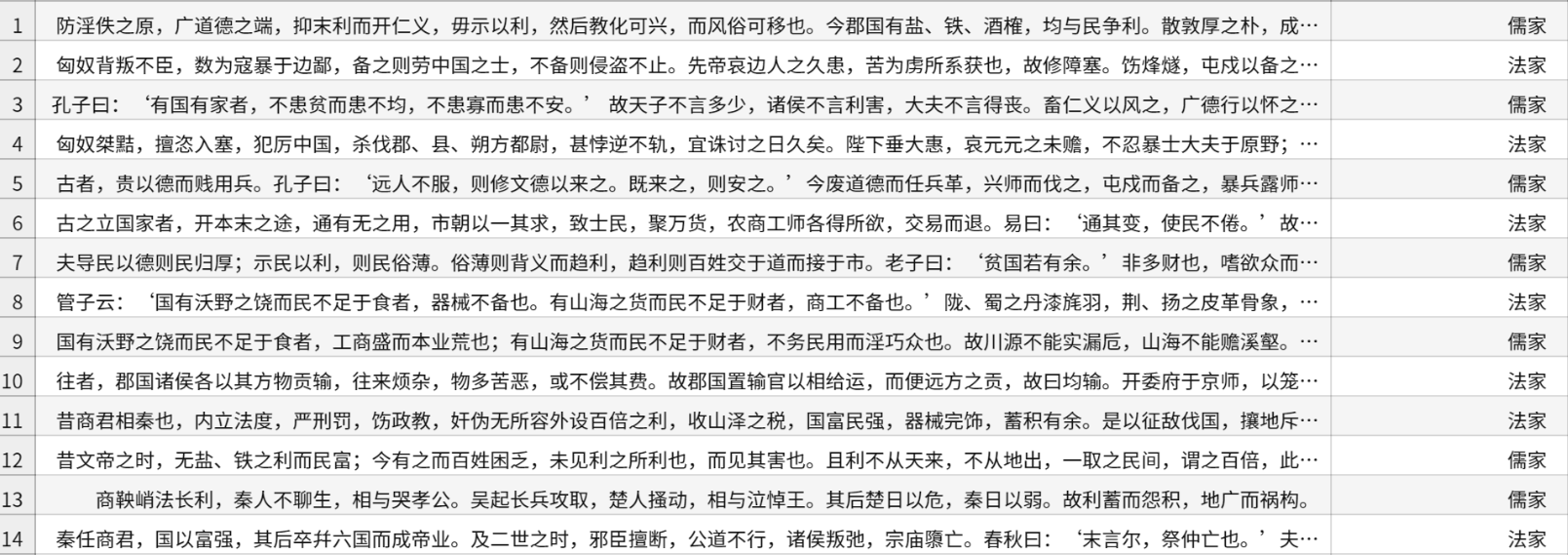}
  \end{subfigure}
  \vspace{-20pt}
  \caption{The Confucianism and Legalism dataset (Chinese version).
}

\end{figure}

\end{document}